\documentclass[sigconf]{acmart}
\usepackage{graphicx}
\usepackage{color}
\usepackage{booktabs}

\AtBeginDocument{%
  \providecommand\BibTeX{{%
    \normalfont B\kern-0.5em{\scshape i\kern-0.25em b}\kern-0.8em\TeX}}}

\copyrightyear{2022}
\acmYear{2022}
\setcopyright{rightsretained}
\acmConference[ICCAD '22]{IEEE/ACM International Conference on Computer-Aided Design}{October 30-November 3, 2022}{San Diego, CA, USA}
\acmBooktitle{IEEE/ACM International Conference on Computer-Aided Design (ICCAD '22), October 30-November 3, 2022, San Diego, CA, USA}
\acmDOI{10.1145/3508352.3549357}
\acmISBN{978-1-4503-9217-4/22/10}

\begin{document}

\title[FastStamp]{FastStamp: Accelerating Neural Steganography and Digital Watermarking of Images on FPGAs}

\author{Shehzeen Hussain$^{1}$\text{*}, Nojan Sheybani$^{1}$\text{*}, Paarth Neekhara$^{2}$\text{*}, Xinqiao Zhang$^1$, Javier Duarte$^3$, Farinaz~Koushanfar$^1$}
\affiliation{%
\department{\text{*}Equal contribution}
  \institution{$^1$Department of Electrical and Computer Engineering, UC San Diego, $^2$Department of Computer Science and Engineering, UC San Diego, $^3$Department of Physics, UC San Diego}
  \city{}
  \state{}
  \country{}
 }
\email{{ssh028,nsheyban, pneekhar,x5zhang,jduarte,fkoushanfar}@ucsd.edu}

\renewcommand{\shortauthors}{Hussain, et al.}

\begin{abstract}
Steganography and digital watermarking are the tasks of hiding recoverable data in image pixels. 
Deep neural network (DNN) based image steganography and watermarking techniques are quickly replacing traditional hand-engineered pipelines. 
DNN based watermarking techniques have drastically improved the message capacity, imperceptibility and robustness of the embedded watermarks. However, this improvement comes at the cost of increased computational overhead of the watermark encoder neural network. 
In this work, we design the first accelerator platform FastStamp to perform DNN based steganography and digital watermarking of images on hardware.
We first propose a parameter efficient DNN model for embedding recoverable bit-strings in image pixels. 
Our proposed model can match the success metrics of prior state-of-the-art DNN based watermarking methods while being significantly faster and lighter in terms of memory footprint. 
We then design an FPGA based accelerator framework to further improve the model throughput and power consumption by leveraging data parallelism and customized computation paths. 
FastStamp allows embedding hardware signatures into images to establish media authenticity and ownership of digital media. 
Our best design achieves $68 \times$ faster inference as compared to GPU implementations of prior DNN based watermark encoder while consuming less power. 


\end{abstract}



\maketitle

\section{Introduction}
Steganography and watermarking techniques aim to embed digital information into visual media in an invisible manner. 
A watermarking scheme typically consists of two components---an \textit{encoder} that embeds a given digital message in the image pixels and a \textit{decoder} that extracts the message from a watermarked image. 
Such technology has several applications, such as transmitting secret messages, copyright protection, establishing media ownership, and embedding invisible QR codes into images~\cite{tancik2020stegastamp}. 
Another important use case of digital watermarking lies in media authentication or integrity verification by embedding semi-fragile watermarks into digital media at the source. 
Semi-fragile watermarks require the property of being fragile to malicious manipulations or tampering while being robust to benign image-processing operations such as image compression, scaling, and color adjustments.

\begin{figure}[h]
\centering
\includegraphics[width=\linewidth]{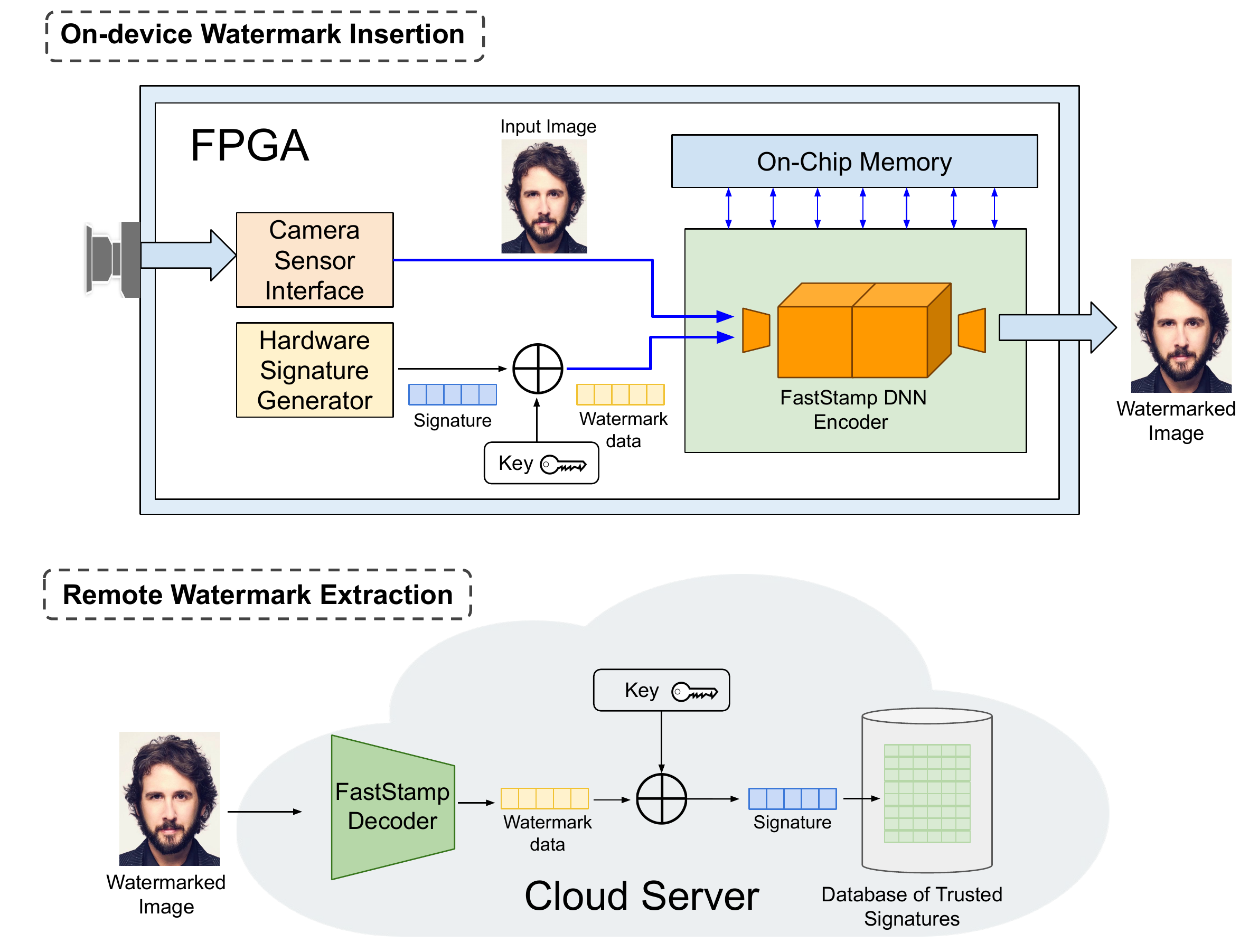}
\caption{Schematic diagram of FastStamp watermarking pipeline. 
The pipeline is divided into two steps: watermark insertion using a DNN encoder on FPGA (top) and watermark extraction using a DNN decoder on a cloud server (bottom).}
\label{fig:overview}
\end{figure}
Conventional image watermarking systems use hand engineered pipelines~\cite{lin2000detection,ho2004semi,yang2009semi,li2015semi} to embed information in the spatial or frequency domain of an image. 
However, the major limitations of traditional approaches lie in the higher visibility of the embedded watermarks, limited message capacity, and low robustness to digital compression techniques like JPEG transforms.
As a result, these traditional approaches are being quickly replaced with deep learning based techniques, which are achieving state-of-the-art results in image watermarking and steganography tasks. 

Deep learning based watermarking techniques relies on encoder and decoder convolutional neural networks (CNNs). 
These networks are trained end-to-end for the task of embedding and retrieving a given message in an image. 
While deep learning techniques significantly outperform hand engineered watermarking pipelines, the improvement comes at the cost of increased computational overhead and memory requirement of these models. 
The best performing neural image watermarking encoders are parameterized by around half a million floating point parameters, which makes it challenging to deploy such systems on resource constrained hardware such as FPGAs or handheld devices.
Such techniques have only been implemented at a software level which results in significant latency between image capture and transmission.
Our work aims to embed watermarks in images and videos in real-time as they are being captured. 
Embedding the watermarks at the hardware level can not only reduce the latency of the watermarking process but also enable media authentication and provenance by leveraging unique hardware signatures from PUFs~\cite{gu2017improved} or secure enclaves as the watermarking data. 

In this work, we propose FastStamp---a light-weight yet robust neural image watermarking framework to enable real-time watermarking on hardware platforms. 
Keeping resource constraints in mind, we develop a parameter efficient CNN based watermarking model that can match and even outperform the success metrics of state-of-the-art neural image watermarking models.
Our watermarking model leverages efficient neural blocks such as depthwise separable convolutions, spatial upsampling operations, and linear layers to reduce the memory requirement and inference latency without compromising the watermark retrieval accuracy, message capacity, and imperceptibility. 
We then design and verify an FPGA implementation of our watermarking encoder model to allow image watermarking directly on hardware. 
Our most optimized design achieves real-time image watermarking and only requires $3$ milliseconds to watermark a given image while achieving the same results as the software implementations. 
To the best of our knowledge, we propose the first framework to accelerate deep neural networks for steganography and watermarking of images using FPGAs. 
FastStamp can be customized for semi-fragile and robust image watermarking. 
While semi-fragile watermarking is useful for checking media integrity, robust image watermarking is useful for establishing media ownership and copyright protection. 

\noindent \textbf{Summary of Contributions:}
\begin{itemize}
    \item We develop a parameter efficient and computationally inexpensive watermarking model that can embed recoverable watermarks at a significantly lower cost as compared to prior neural watermarking techniques. 
    \item We develop the first FPGA accelerator platform for DNN based image watermarking and steganography. 
    We design reusable and reconfigurable basic blocks pertinent to such models as separable convolutions, 2D upsampling, and skip connections. 
    We deploy the FastStamp encoder model on Xilinx XCVU13P FPGA, which achieves $68 \times$ faster inference than prior neural watermarking models. 
    
    \item Our framework is end-to-end and supports two types of watermarking schemes: \textit{robust} and \textit{semi-fragile}. FastStamp learns to be robust to a wide range of real-world digital image processing operations such as lighting, color adjustments, and compression techniques. 
    Our semi-fragile watermarking scheme learns to be robust to above benign transformations while being fragile to media forgery and local tampering. 
\end{itemize}


\vspace{-2mm}
\section{Background}
\subsection{Digital Watermarking}
%
Digital watermarking techniques broadly seek to generate three different types of watermarks: fragile~\cite{di2019fragile}, robust~,\cite{cox1997secure,shehab2018secure,zhu2018hidden} and semi-fragile~\cite{lin2000detection,yu2017review,ho2004semi}.
Fragile and semi-fragile watermarks are primarily used to certify the integrity and authenticity of image data.
Fragile watermarks are used to achieve accurate authentication of digital media, where even a one-bit change to an image will lead it to fail the certification system.
In contrast, \textit{robust} watermarks aim to be recoverable under several image manipulations to allow media producers to assert ownership over their content even if the video is redistributed and modified.
Semi-fragile watermarks combine the advantages of both robust and fragile watermarks and are mainly used for fuzzy authentication of digital images and identification of image tampering~\cite{yu2017review}. 

Prior work has proposed hand-engineered pipelines to embed semi-fragile watermark information in the spatial\cite{xiao2008semi} and frequency domain of images and videos. 
Particularly, in the frequency domain, the watermark can be embedded by modifying the coefficients produced with transformations such as the discrete cosine transform (DCT)~\cite{preda2015watermarking,ho2004semi} and discrete wavelet transform (DWT)~\cite{tay2002image,li2015semi,benrhouma2015tamper}. 
For example, during 2D DWT an image is first decomposed into various frequency channels using a Haar filter. 
A scaled image is used as the watermark and inserted into mid frequency wavelet channel. 
Taking 2D inverse DWT of the altered wavelet decomposition produces the watermark embedded image. 
The major drawbacks of traditional approaches lie in higher visibility of the embedded watermarks, increased distortions in generated images, and low robustness to compression techniques like JPEG transforms as discussed in Section~\ref{sec:evaluation_metrics}. 
Specifically, since digital images shared over social media are highly compressed and undergo various lighting and color adjustments, the watermarks generated by DWT and DCT algorithms break under benign real-world transformations and compression.

More recently, CNNs have been used to provide an end-to-end solution to the watermarking problem. They replace hand-crafted hiding procedures with neural network encoding~\cite{baluja2017hiding,hayes2017generating,zhu2018hidden,zhang2019invisible,wang2021faketagger,tancik2020stegastamp,neekhara2022facesigns,luo2020distortion}. 
These techniques train end-to-end encoder CNNs to embed and decode watermarks, which have resulted in lower imperceptibility and more robust recovery of the watermark data. 
However, these models are memory intensive and much slower as compared to DCT based algorithms. 
In our work, we address the performance limitations of neural watermarking systems and propose a light-weight framework for both robust and semi-fragile image watermarking suitable for hardware acceleration platforms.

\subsection{FPGA Accelerated Techniques}

There have been several efforts to accelerate neural networks on FPGAs~\cite{zhang2015optimizing,suda2016throughput,samragh2017customizing,fastwavepaper}. Equipped with the necessary hardware for basic DNN operations, FPGAs are able to achieve high parallelism and utilize the properties of neural network computation to remove unnecessary logic. 
Some prior efforts have been made in FPGA acceleration of convolutional autoencoder architectures~\cite{liu2018optimizing,zhao2018fpga,govorkova2022autoencoders}. 
While these works implement many sub-blocks used in neural network computation, we find that they cannot be directly used for an image watermarking framework because efficient implementations of sub-blocks like 2D upsampling and separable convolutions with skip-connections are absent. Moreover, existing neural architectures for image watermarking are not optimized for hardware-software co-design. 

Prior work~\cite{kiran2013design,hazra2018fpga,hajjaji2019fpga,khoshki2014fpga} has made significant efforts in accelerating traditional image watermarking schemes that hide secret information in the frequency domain of images and rely on DCT and DWT based algorithms.
As discussed earlier, while such algorithms are vastly popular for hardware applications due to its simplicity and low computational overhead, the resulting watermarked image is often not robust to real-world image transformations. In our work, we develop the first FPGA accelerator platform for DNN based image watermarking and steganography that enables robustness to real-world digital image processing and compression while being selectively fragile to media tampering techniques. 

\subsection{Countering Media Forgery}
With the widespread development of deep learning based image and video synthesis techniques~\cite{liu2019stgan,choi2020starganv2,dfsurvey,faceswap}, it has become increasingly easier and faster to generate high-quality convincing fake images/videos such as Deepfakes. 
Such manipulated media can fuel misinformation, defame individuals and reduce trust in media. 
While considerable research effort has been made in designing CNN based deepfake detectors~\cite{dolhansky2020deepfake,afchar2018mesonet}, these techniques have been shown to hold major security vulnerabilities and can be bypassed by attackers~\cite{advdeepfakes}.
To counter such threats, authors~\cite{9689555,neekhara2022facesigns,wang2021faketagger} have proposed semi-fragile watermarking as a solution to perform media authentication and distinguish deepfake media from real media by verifying a secret watermark embedded in the media. 
For both fragile and semi-fragile techniques, a watermark must be inserted when the image is captured, which makes these techniques dependent on both algorithmic and hardware implementation. 
If watermark information is embedded separately in images and videos after it is captured by a device, this method may fail in situations where tampering is carried out before inserting the signature or watermark. 
While the proposed solution to media authentication is to add a verifiable digital signature or watermark to an image/video using neural networks, prior works~\cite{neekhara2022facesigns,9689555,wang2021faketagger} do not actively implement the technology in resource constrained settings or camera hardware. 
Upon empirical study, we found that it is challenging to fit such off-the-shelf models~\cite{tancik2020stegastamp,zhu2018hidden,neekhara2022facesigns} on FPGAs since they were developed without any attention to hardware-software co-design practices and range from 500\,k to 2 million parameters for the encoder model. 
To this end, we design our own DNN based watermarking system that utilizes depthwise separable convolutions and a parameter efficient message upsampler to reduce computational overhead while preserving required bit recovery accuracy, capacity, and imperceptibility. 

\begin{figure*}
    \centering
    \includegraphics[width=1.0\linewidth]{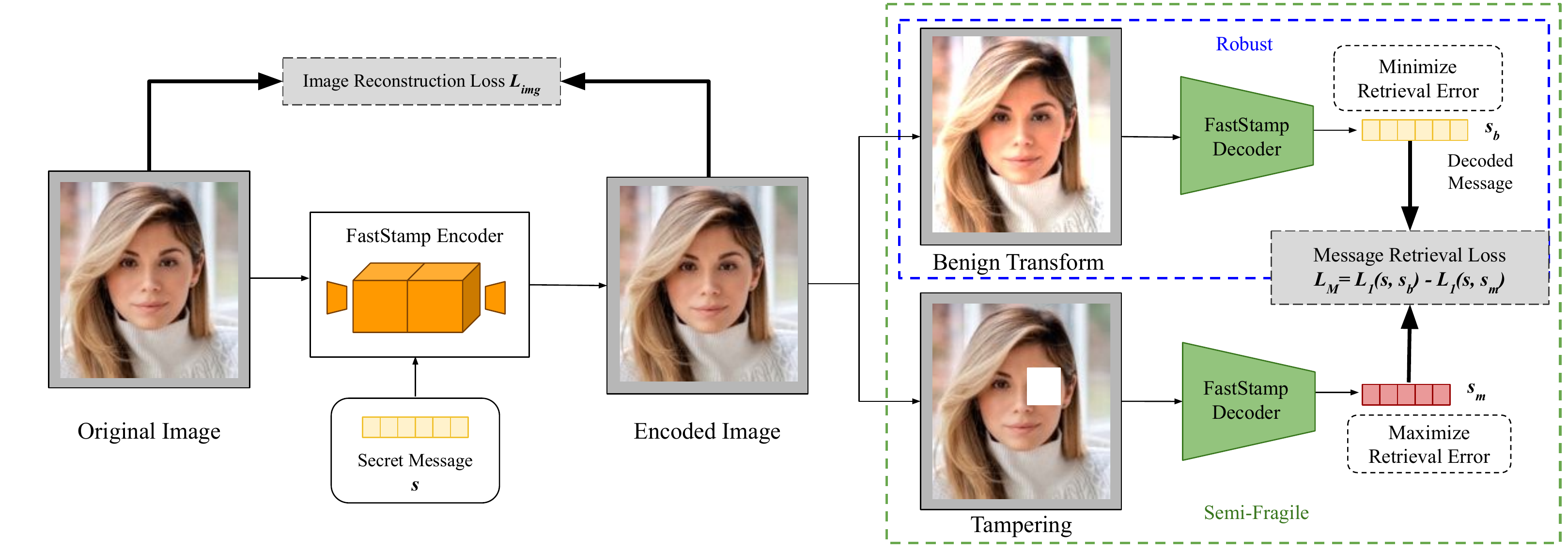}
    \caption{FastStamp Encoder-Decoder Training framework. 
    In both robust and semi-fragile schemes, the encoder model learns to embed a message in a given image by encouraging retrieval from the decoder model under benign transforms. 
    For our semi-fragile design, we additionally encourage fragility to malicious transforms by maximizing message retrieval error on tampered watermark images.}
    \label{fig:model_diagram}
\end{figure*}

\section{Methodology}
\subsection{Training Framework}
\label{sec:training}
Our goal is to develop a learnable, end-to-end model for image steganography and watermarking such that the encoder model can embed a message as a visually invisible perturbation in the image, and the decoder network can extract the message from the watermarked image. 
We develop two variants of our training framework to generate \textit{robust} and \textit{semi-fragile} watermarks. 
A robust watermark is designed to be recoverable when real-world image transformations are applied. 
A semi-fragile watermark is designed to be robust to benign image transformations such as compression, minor color, and contrast adjustments but it should be unrecoverable when malicious image transforms such as image tampering and face-swapping are applied. 
A semi-fragile watermark is designed to be robust to benign image transformations such as compression and minor color, and contrast adjustments but it should be unrecoverable when malicious image transforms such as image tampering and face-swapping are applied. 

The encoder network $E$ takes as input an image $x$ and a bit string $s \in \{0, 1\}^L$ of length $L$, and produces an encoded (watermarked) image $x_w$. 
That is, $x_w=E(x, s)$. 
Depending on our task of either \textit{semi-fragile} or \textit{robust} watermarking, the encoded image goes through the following operations: 
\begin{enumerate}
\item \textit{\textbf{Robust watermarking:}}
In this setting, the image goes through a benign image transformation $g_b \sim G_b$ to produce $x_b=g_b(x_w)$. 
The benign image is then fed to the decoder network, which predicts the message $s_b=D(x_b)$. 
For optimizing secret retrieval during training, we use the $L_1$ distortion between the predicted and ground-truth bit strings. 
The decoder is encouraged to be robust to benign transformations by minimizing the message distortion $L_1(s, s_b)$.  
Therefore the secret retrieval error for an image $L_M(x)$ is obtained as follows:
    \begin{equation}
    L_M(x) = L_1(s, s_b)
    \end{equation}
\item \textit{\textbf{Semi-fragile watermarking:}} In this setting, the watermarked image goes through two image transformation functions---one sampled from a set of benign transformations ($g_b \sim G_b$) and the other sampled from a set of malicious transformations  ($g_m \sim G_m$) to produce a benign image $x_b=g_b(x_w)$ and a malicious image $x_m=g_m(x_w)$. 
The benign and malicious watermarked images are then fed to the decoder network, which predicts the messages $s_b=D(x_b)$ and $s_m=D(x_m)$ respectively. 
The decoder is encouraged to be robust to benign transformations by minimizing the message distortion $L_1(s, s_b)$; and fragile for malicious manipulations by maximizing the error $L_1(s, s_m)$. 
Therefore the secret retrieval error for an image $L_M(x)$ is obtained as follows:
    \begin{equation}
    L_M(x) = L_1(s, s_b) - L_1(s, s_m)
    \end{equation}
\end{enumerate}

The watermarked image is encouraged to look visually similar to the original image by optimizing three image distortion metrics: $L_1$, $L_2$ and $L_\text{pips}$~\cite{zhang2018unreasonable} distortions. 
\begin{equation}
\begin{split}
& L_\text{img}(x, x_w) = L_1(x, x_w) + L_2(x, x_w) + c_p L_\text{pips}(x, x_w)
\end{split}
\end{equation}

Therefore, the parameters $\alpha,\beta$ of the encoder and decoder network are trained using mini-batch gradient descent to optimize the following loss over a distribution of input messages and images: 
\begin{equation}
\mathbb{E}_{x, s, g_b, g_m} [ L_\text{img}(x, x_w) + c_M L_M(x)  ]
\end{equation}

In the above equations, $c_p$, and $c_M$ are scalar coefficients for the respective loss terms. 
We refer the readers to our supplementary material for the values we use for these coefficients and other implementation details.

\subsection{Message encoding} The input of our encoder network is a bit string $s$ of length $L$. This watermarking data includes a secret message or a hardware signature generated by trusted execution environments or PUFs. 
To further ensure message secrecy, we can encrypt the message using a stream cipher with a secret key that is shared between the encoding and decoding devices. 
In our work we embed messages of length 128 bits in an image of size $128{\times}128$, which allows embedding $2^{128}$ unique messages in each $128{\times}128$ image patch. 



\subsection{Model Architecture and Optimization}
\label{sec:architecture}
The encoder model takes as input an image $x$ and a message bit-string $s$ to produce a watermarked image $x_w$.
The encoder model of a typical neural watermarking system follows a convolutional U-Net architecture comprising several downsampling and upsampling layers with skip-connections. 
In prior work~\cite{tancik2020stegastamp,neekhara2022facesigns}, the secret message bit-string is first projected using a learnable linear layer reshaped as a matrix to have the same height and width as the input image; and then attached as the fourth channel of the input image. 
The combined input and secret image then undergo the downsampling and upsampling operations of the U-Net to produce the watermarked image.

The above described encoder model architecture in prior work~\cite{tancik2020stegastamp,zhu2018hidden,neekhara2022facesigns} has a large memory footprint and is unsuitable for deployment in resource-constrained settings. To reduce the model size without compromising on the watermarking performance, we propose the following architectural optimizations:

\subsubsection{\textbf{Secret Message Upsampler}}
The secret message upsampler projects the message string $s$ to a matrix that gets attached as the fourth channel of the input image. 
A na\"{i}ve implementation using a linear layer can result in a large memory footprint since the number of parameters is given by $hwL$, where $h$ and $w$ are the input image height and width, respectively, and $L$ is the secret message length. 
An input image size of $128{\times}128$ and a message length of $128$, would result in more than two million parameters.
To optimize the number of parameters, we perform the secret upsampling operation as follows:
\begin{enumerate}
    \item Project the message $s$ to a vector $s_\text{proj}$ of size $h'w'$ using a linear layer.
    \item Reshape $s_\text{proj}$ to a matrix $s_\text{proj}{}_M$ of dimensions $(h', w')$
    \item Upsample $s_\text{proj}{}_M$ to a matrix of size $(h, w)$ using nearest-neighbour upsampling.
\end{enumerate}

\begin{figure}[htp]
\centering
\includegraphics[width=1.0\linewidth]{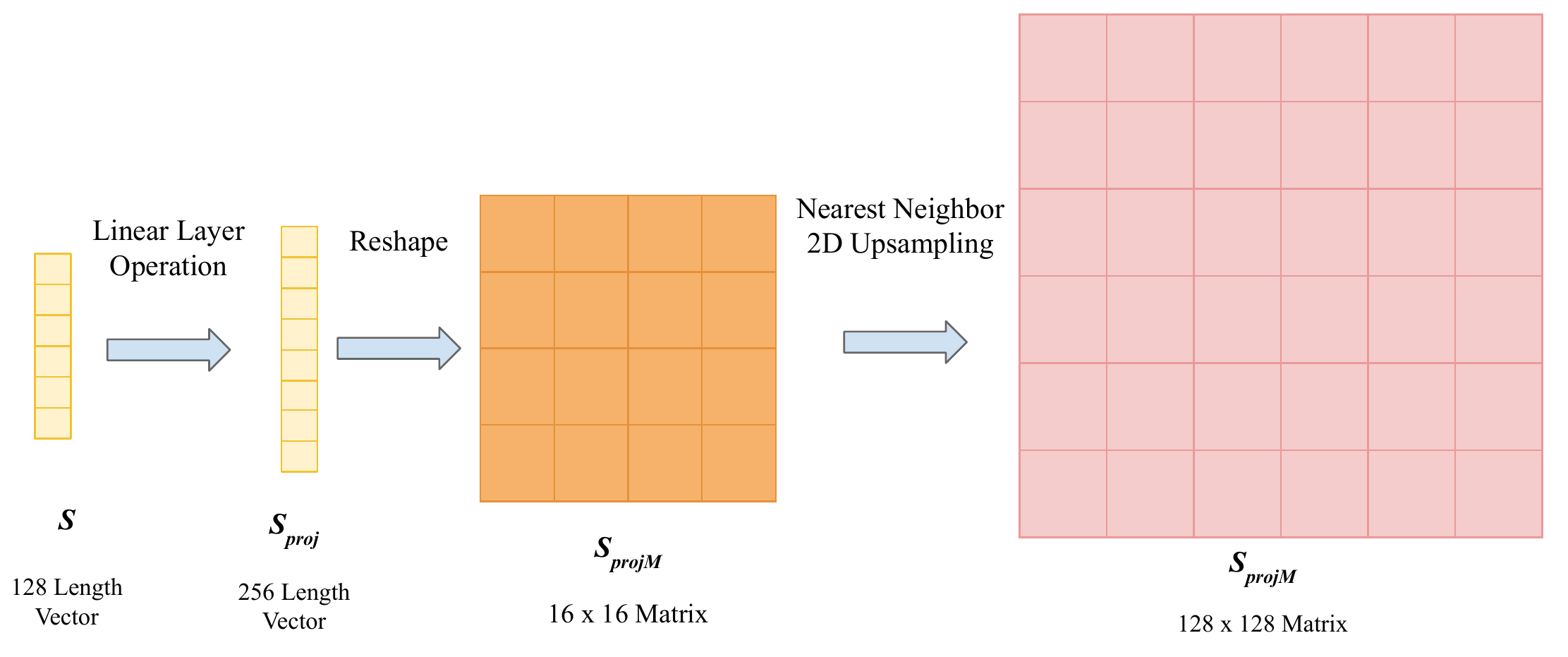}
\caption{An example of optimized secret message upsampling using linear layer projection followed by nearest neighbor 2D upsampling.}
\label{fig:overview}
\end{figure}

The nearest-neighbour upsampling operation is parameter-free and computationally more efficient as compared to matrix-vector multiplication. 
Through our experiments, we find that using an $(h', w')$ that are much smaller than $(h, w)$ can achieve the same watermarking while being significantly efficient in both time and memory. 
For an image of size $(128, 128)$ and message length $L=128$, we use $h'=w'=16$ thus requiring only $32768$ parameters. 
The upsampled secret gets attached as the fourth channel of the input image and undergoes the U-Net downsampling and upsampling operations described below. 

\subsubsection{\textbf{U-Net Downsampling}}

\begin{figure}[htp]
\centering
\includegraphics[width=\linewidth]{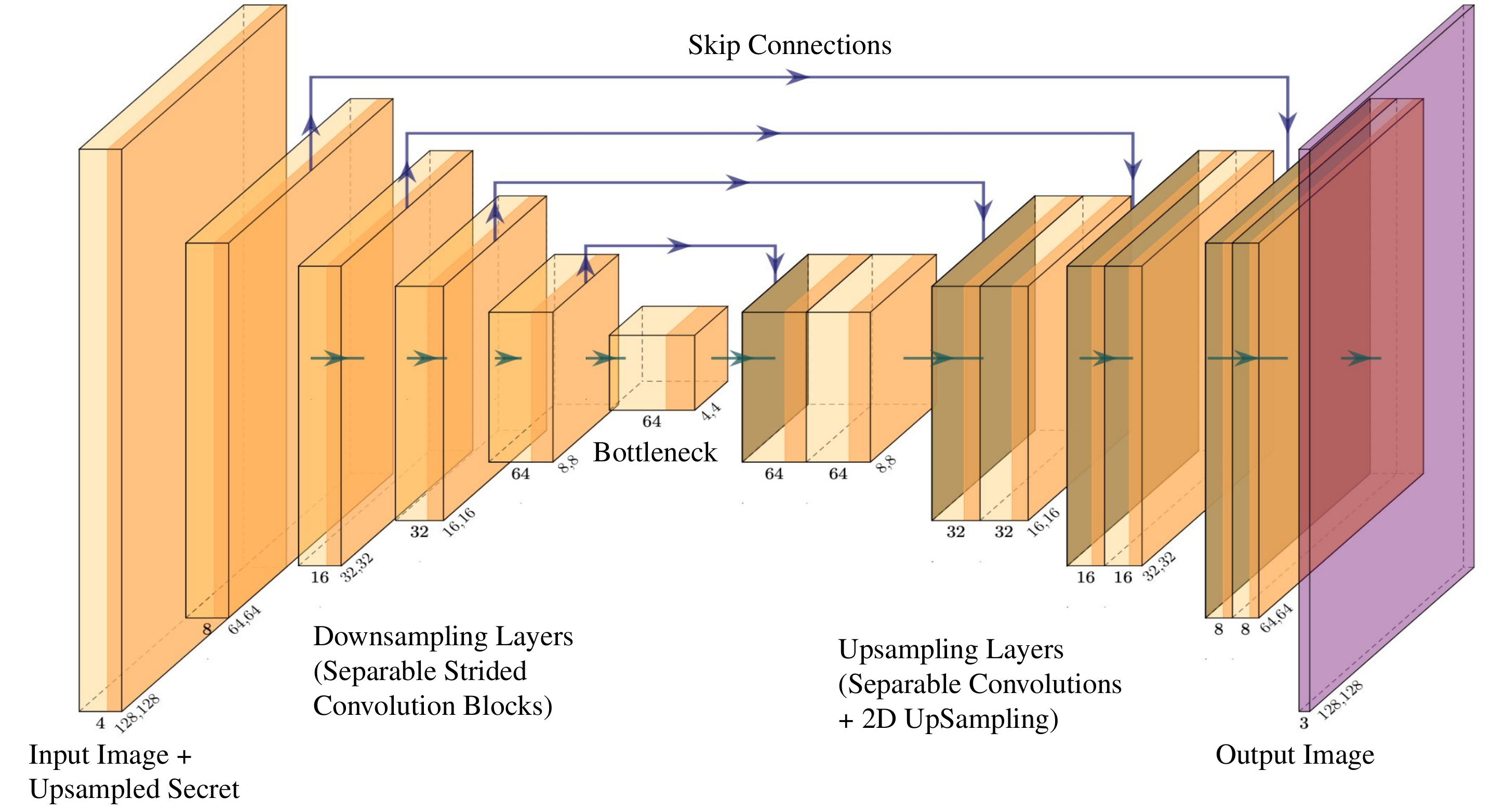}
\caption{FastStamp Encoder Architecture. 
Our encoder network takes as input an image $x$ and the output of the secret message upsampler $s_\text{proj}{}_M$ and generates the watermarked image}
\label{fig:unet}
\end{figure}

The downsampling network in U-Net architecture typically comprises $5$ to $8$ convolutional blocks. 
Each block contains a strided convolutional layer, a batch normalization layer, and a non-linear activation like ReLU or leaky ReLU. 
The number of output channels of each convolutional layer increase with the depth of the network, doubling at each step until a maximum value is reached. 
To optimize this architecture, we first replace the convolutional layers with depthwise and point-wise separable convolutional layers~\cite{depthwise}. 
Not only does this optimization reduce the number of parameters but also reduces the number of floating point operations required in each layer computation. 
Next, we optimize a number of output channels of each convolutional block. 
In our experiments, we perform a design-space exploration to find that we can substantially reduce the number of output channels in each layer without compromising on secret retrieval accuracy and imperceptibility. 
Our most optimized design uses $5$ downsampling layers with $64$ output channels in the final separable convolutional layer. 
Figure~\ref{fig:unet} details the network architecture and output tensor sizes after each downsampling step.

\subsubsection{\textbf{U-Net Upsampling}}
The upsampling network in U-Net architecture follows a mirror image of the downsampling network. 
Instead of a regular convolutional layer, each upsampling block typically contains a transposed convolutional layer. However, transposed convolutional layers have been shown to introduce unwanted visual artifacts in the generated images~\cite{odena2016deconvolution} and we see the same effect empirically in our work.
To remove such artifacts, we replace the transposed convolution layer with a separable convolution layer followed up by nearest neighbor 2D upsampling following the recommendations given by past work~\cite{odena2016deconvolution}.
At each upsampling step, the output of the corresponding downsampling step is concatenated with the block input to provide skip-connections which are known to improve the performance of encoder-decoder models.
The output of the last upsampling layer undergoes a $\tanh$ activation function to normalize output values between $-1$ and $1$, which are then scaled between $0$ and $1$ to produce an RGB image. 
Figure~\ref{fig:unet} details our encoder network architecture.

FastStamp's Decoder follows a similar architecture as the encoder but contains 8 downsampling and 8 upsampling layers. After the upsampling U-net, the decoder network follows the inverse architecture of the secret message upsampler to predict the 128-bit message. 

\section{Accelerator Design}

\subsection{Design Overview}

Figure~\ref{fig:overview} gives a high-level overview of our FPGA accelerator design. 
Each neural network layer is treated as a separate dataflow stage. 
To be low latency and high throughput, all weights and biases are stored on-chip.
Complex activation functions are implemented via precomputed lookup tables. 
The design uses task-level pipelining (i.e., HLS dataflow) for each layer and streams the data between each dataflow stage using first-in-first-out buffers (FIFOs).
As FIFOs can only be read once, to implement the skip connections, an additional dataflow stage is used to clone the skip connection data from its input FIFO into two other FIFOs so that it can be read twice for its two datapaths.


\begin{figure}[htp]
\centering
\includegraphics[width=\linewidth]{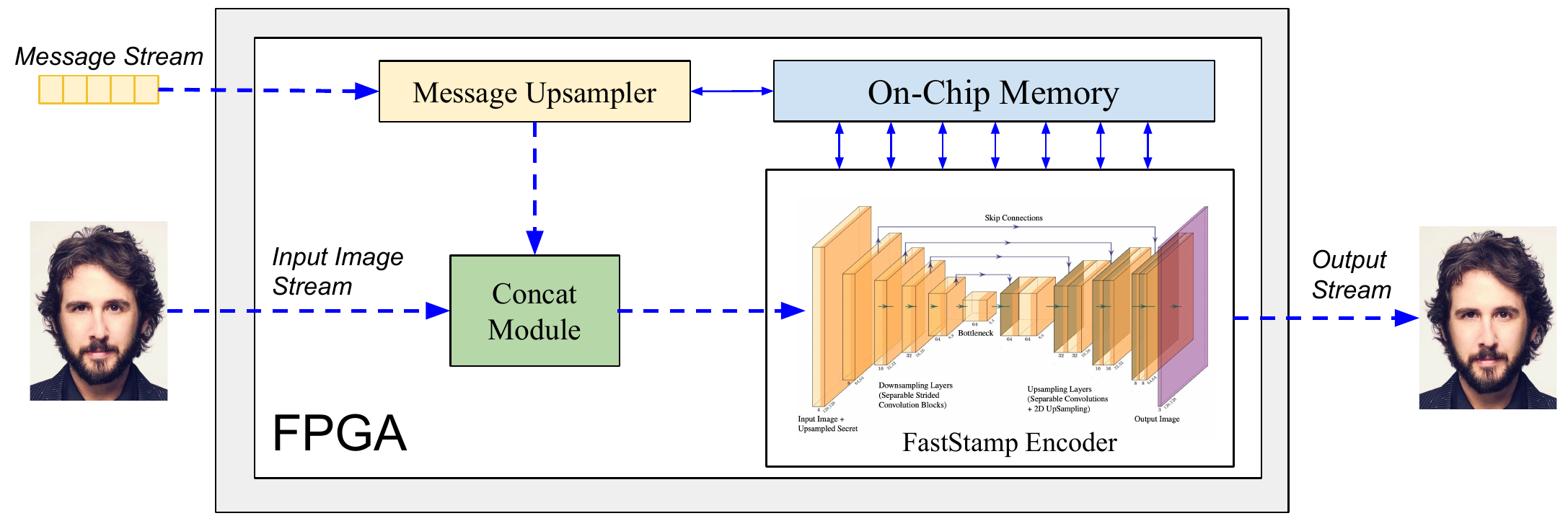}
\vspace{-6mm}
\caption{Design overview of FastStamp Accelerator Platform.}
\label{fig:overview}
\end{figure}

\vspace{-4mm}
\subsection{Implementation Details}
\label{sec:implementation}
To implement FastStamp on FPGA, we began with using the standard \textit{hls4ml}~\cite{Duarte:2018ite,Aarrestad:2021zos,Fahim:2021cic} framework and added additional libraries to this to support the necessary functionalities such as depthwise separable convolutions, and nearest-neighbor 2D upsampling, concatenation layers (i.e., skip connections) which were previously not supported. 
We create custom sub-blocks and modify existing implementations with optimizations that better support our model. 
Our design is composed of the following $6$ main architectural sub-blocks:

\subsubsection{\textbf{Linear Layer}}
The linear layer in the \textit{Secret Message Upsampler} module is implemented using efficient matrix-vector multiplication based on product and sum tree. 
In order to optimize the design and make efficient use of the DSP blocks, we use a parallelized approach to convert layer computations into multiple MAC operations. 
Multiple rows of the weight matrix are processed simultaneously by dividing it into chunks. 
In each round, a chunk of the weights matrix is copied to one of the weight buffers while the other weight buffer is fed into the \textit{dot product} modules together with a copy of the input vector. 
The iterations end when all rows of the weight matrix have been processed. 
Then each \textit{dot-product} function partitions its input vectors into chunks and concurrently executes MAC operations over the partitioned subsets. 
The accumulated results of the subsets are then added together within the \textit{reduce\_sum} function to compute the final output. 
The \textit{reduce\_sum} module performs a tree-based reduction algorithm which further reduces latency. 

\subsubsection{\textbf{Nearest-neighbor 2D Upsampling}}
2D upsampling layers are used in the secret \textit{Secret Message Upsampler} and the \textit{U-Net Upsampling layers}. 
To upsample a matrix, we iterate through every element in the matrix and start building the new resized matrix. 
We use nearest neighbour as our interpolation method, as it is the least hardware intensive computation available for resizing purposes. 
We do not utilize pipelining in this sub-block, as it would cause a drastic increase in resource utilization for this resizing task while not having a substantial effect on latency. 
Due to these problems, pipelining makes this sub-block unscalable. Instead, loop unrolling is utilized to reduce loop iterations, giving us latency reductions with negligible impact on resource utilization.

\subsubsection{\textbf{Skip Connections}}
Concatenate layers are used in FastStamp to implement skip connections between upsampling and downsampling layers. 
These layers contributed towards some of the highest resource utilization in our design. 
To realize FastStamp on hardware, efficient support for concatenation of different sized inputs was implemented. 
We also distribute resource utilization to BRAM within these layers due to a large allocation of LUTs. 
Alongside this, we are able to minimize latency by pipelining the operations with complete unrolling in these layers. 

\subsubsection{\textbf{Separable Convolution}}
Due to resource limitations, traditional convolution techniques in machine learning settings are often not feasible on hardware. 
Recent works~\cite{bai2018sepconv, yoo2018cnn} have shown that utilizing separable convolutions significantly reduces the number of multiplications needed. 
Instead of traditional convolution,  separable convolution layers are used in both U-Net downsampling and upsampling layers. 
A separable convolution comprises of a depthwise separable convolution and point-wise convolution operation. 
The depthwise separable convolution is performed independently across the input channels and results in the same number of output channels. 
A point-wise convolution operation reduces to matrix-vector multiplication along the channel axis for each spatial cell of the input. 
We also use a streaming implementation of depthwise convolution utilizing pipelining with complete loop unrolling, array partitioning, and loop flattening to achieve minimal latency. 
We optimize standard point-wise convolution in a similar manner as our linear layer. 

\subsubsection{\textbf{Batch-normalization}}
Batch-normalization~\cite{pmlr-v37-ioffe15} layers are used in U-Net downsampling and upsampling layers. 
These layers are used after the separable convolution layers to stabilize the network. 
Rather than using this optimization, pipelining, loop unrolling, and array partitioning are enforced to accelerate the normalization layer. 
This allows for the batch-normalization output to be used as an input to skip connections, a feature that cannot be done when fusing layers. 
Our optimizations achieve low latency and minimal resource utilization.

\subsubsection{\textbf{Non-linear activations}}
ReLU and $\tanh$ are the two non-linear activation functions are used in FastStamp.
The ReLU function, which is used in the U-net downsampling and upsampling layers computes the following function for each input $x$, $y = \max(0, x)$. 
Due to the simplicity of the ReLU activation function, we use loop unrolling to optimize latency for this operation. 
The $\tanh$ activation function computes the following function for each input $x$, $y=(e^x - e^{-x})/(e^x + e^{-x})$. 
We use the default $\tanh$ implementation in \textit{hls4ml} which is performed using a pre-computed lookup table to reduce latency.

\section{Experiments and Results}

\setlength\tabcolsep{4pt}
\begin{table*}[htp]
\centering
\begin{tabular}{@{}l|r|rrr|cc|ccc|c@{}}
\multicolumn{1}{r}{} & \multicolumn{1}{|r|}{\emph{Model Size}} & \multicolumn{3}{c|}{\emph{Capacity}} & \multicolumn{2}{c|}{\emph{Imperceptibility}} & 
\multicolumn{3}{c|}{\emph{BRA (\%) -- Benign}} &
\multicolumn{1}{c}{\emph{BRA (\%) -- Tampering}}
\\
\midrule
Method & \# Params & $H$,$W$ & $L$ & $\text{BPP}$ & PSNR & SSIM & None & JPG-75 & Filtering & FaceSwap \\ 
\midrule

DCT (Semi-Fragile)~\cite{ho2004semi} & --- & $128$ & $256$ & $5.2\times10^{-3}$ & $22.49$ & $0.871$ & $99.81$ & $56.65$ & $94.62$ & $85.51$ \\
HiDDeN (Robust)~\cite{zhu2018hidden} & 411\,k & $128$ & $30$ & $6.1\times10^{-4}$ & $27.57$ & $0.934$ & $97.06$ & $72.71$ & $94.52$ & --- \\
StegaStamp (Robust)~\cite{tancik2020stegastamp} & 528\,k & $400$ & $100$ & $2.0\times10^{-4}$  & $29.39$ & $0.925$ & $99.92$ & $99.91$ & $99.84$ & --- \\
\midrule
FastStamp (Robust) & 45\,k & $128$ & $128$ & $2.3\times10^{-3}$ & $30.65$ & $0.942$ & $100.00$ & $99.84$ & $99.78$ & --- \\
FastStamp (Semi-Fragile) & 45\,k & $128$ & $128$ & $2.3\times10^{-3}$ & $30.64$ & $0.940$ & $100.00$ & $99.74$ & $99.72$ & $51.11$ \\

\bottomrule
\multicolumn{1}{c}{}
\end{tabular}
\vspace{-2mm}
\caption{Capacity, imperceptibility, and BRA metrics of different watermarking systems. $H,W$ indicates the height and width of the input image. For both robust and semi-fragile watermarking systems a high BRA is desirable for benign transforms. For semi-fragile watermarking systems, a low BRA is desirable for tampering transforms.}

\label{tab:basicompare}
\end{table*}

\subsection{Dataset} 
We conduct experiments on the CelebA dataset~\cite{liu2015faceattributes} which is a large database of over 200,000 face images of 10,000 unique celebrities. 
We set aside 1000 images for testing the watermarking models and split the remaining data into 80\% training and 20\% validation.
All FastStamp models are trained using images of size $128{\times}128$, which are obtained after center-cropping and resizing the CelebA images.
We conduct experiments with message bit length $L=128$.

\subsection{Evaluation Metrics}
\label{sec:evaluation_metrics}
For the evaluation of our watermarking techniques, we investigate the following metrics based on prior works~\cite{hore2010image,hajjaji2019fpga}.

\noindent \textbf{1. Imperceptibility:} We compute \textbf{peak signal to noise ratio (PSNR)} and \textbf{structural similarity index (SSIM)} between the watermarked and original images. 
A higher value of both these metrics indicates a more imperceptible watermark.

\noindent \textbf{2. Capacity:} Capacity measures the amount of information that can be embedded in the image. 
We use \textbf{bits per pixel (BPP)} which is calculated as $L/(HWC)$ where $L$ is the message length and $H,W,C$ indicate the height, width, and channels of the image. 
Higher BPP values indicate higher capacity.

\noindent \textbf{3. Bit-Recovery Accuracy 
(BRA):} BRA calculates the recovery accuracy of the bit string $s$. 
For robustness, we aim to have a high BRA when benign or intended transformations such as JPEG compression or color and contrast adjustments are applied. 
For semi-fragile watermarking systems, the goal is to have a low BRA when image tampering operations such as local tampering or FaceSwap are applied while maintaining robustness against benign transformations. 

We compare our watermarking framework against three prior works on image watermarking---a DCT based semi-fragile watermarking system~\cite{yang2009semi} and two robust neural image watermarking systems HiDDeN~\cite{zhu2018hidden} and StegaStamp~\cite{tancik2020stegastamp}. 
To evaluate the robustness and fragility of our watermarking systems, we perform the following transformations that are unseen during training:
\begin{enumerate}
\item
\textbf{JPEG Compression: }
Digital images are usually stored in a lossy format such as JPEG. 
We compress the watermarked images using JPEG-75 compression and measure the decoding BRA.
\item
\textbf{Filtering:}
We apply a set of real-world image filtering operations using the Pilgram library~\cite{pilgramfilter} that simulates photo editing filters that are common on social media. These include color, contrast, and lighting adjustments.
\item
\textbf{Face Swapping: }
For evaluating semi-fragile watermarking systems, we simulate image tampering by performing face swapping using the open source implementation of FaceSwap~\cite{faceswap}. 
A low BRA is desirable against this transform to detect tampering.
\end{enumerate}

\vspace{-2mm}
\subsection{Training and Architecture Optimization}
Our encoder model follows the depthwise separable convolutional U-Net architecture as discussed in Section~\ref{sec:architecture}. 
To find the optimum architecture, we create different-sized versions of the baseline U-Net architecture by reducing the number of channels in each layer by a factor of 2, 4 and 8.  
We find that reducing the number of channels by a factor of 4 does not compromise model performance while being significantly lighter than the base U-Net model. 

We train two variants of this optimized design in the robust and semi-fragile settings using the training technique described in Section~\ref{sec:training}. 
In the robust setting, we simulate differentiable JPEG compression, Gaussian blur, color, and contrast adjustment as the benign transforms $g_b$ during training. 
In the semi-fragile setting, in addition to the benign transforms, we simulate differentiable localized tampering as the malicious transform $g_m$ during training. 
We train our models for 200\,k mini-batch iteration with a fixed learning rate of $1.5\times 10^{-4}$ using Adam optimizer. 
Table~\ref{tab:basicompare} compares our optimized models \textit{ FastStamp (Robust) } and \textit{ FastStamp (Semi-Fragile) } against prior neural and DCT based image watermarking frameworks. 
As compared to prior neural image watermarking and steganography models, FastStamp is significantly smaller and achieves a similar BRA with slightly improved imperceptibility as compared to StegaStamp~\cite{tancik2020stegastamp} and HiDDeN~\cite{zhu2018hidden}. 
The improvement in imperceptible metrics is achieved by using nearest neighbor 2D upsampling in the U-Net architecture as opposed to transposed convolutions in prior work.

\setlength\tabcolsep{3pt}
\begin{table*}[htp]
\centering
\begin{tabular}{@{}l|rrrr|ccc|rrr@{}}
\multicolumn{1}{r}{} &
\multicolumn{4}{c|}{\emph{Resource Utilization (\%)}} &
\multicolumn{3}{c|}{\emph{Performance}} & \multicolumn{3}{c}{\emph{Correctness}}
\\
\midrule
  Design & BRAM & FF & LUT & DSP & Clock&  Latency & Throughput & BRA & PSNR & SSIM \\
   \textit{\small{Available Resources}} & 94 Mb & 3456K & 1728K & 12288 & Period (ns)&   (\# Cycles) & (Hz) &  &  &  \\
  \midrule
FixedPoint-32 & $>$100\% & 51\% & $>$100\% & $>$100\% & --- & --- & --- & 100.0 & 30.67 & 0.942\\
FixedPoint-16 & 89\%  & 18\% & $>$100\% & 54\% & --- & --- & ---  & 100.0 & 30.64 & 0.941\\
\textbf{FixedPoint-16-Optimized} & 59\% & 14\% & 72\% & 53\% & 5 & 596823 & 335 & 100.0 & 30.64 & 0.941\\
\bottomrule
\multicolumn{1}{c}{}
\end{tabular}
\caption{Design-space exploration for FPGA implementation of FastStamp on Xilinx XCVU13P FPGA board. 
We report the resource utilization, timing performance, and implementation correctness for various designs. 
Our optimized 16-bit fixed point implementations fit within the available resources while maintaining the same correctness metrics as the 32-bit implementation.}
\label{tab:fpga}
\end{table*}



\vspace{-2mm}
\subsection{Design Space Exploration}
\begin{figure}[htp]
\centering
\includegraphics[width=\linewidth]{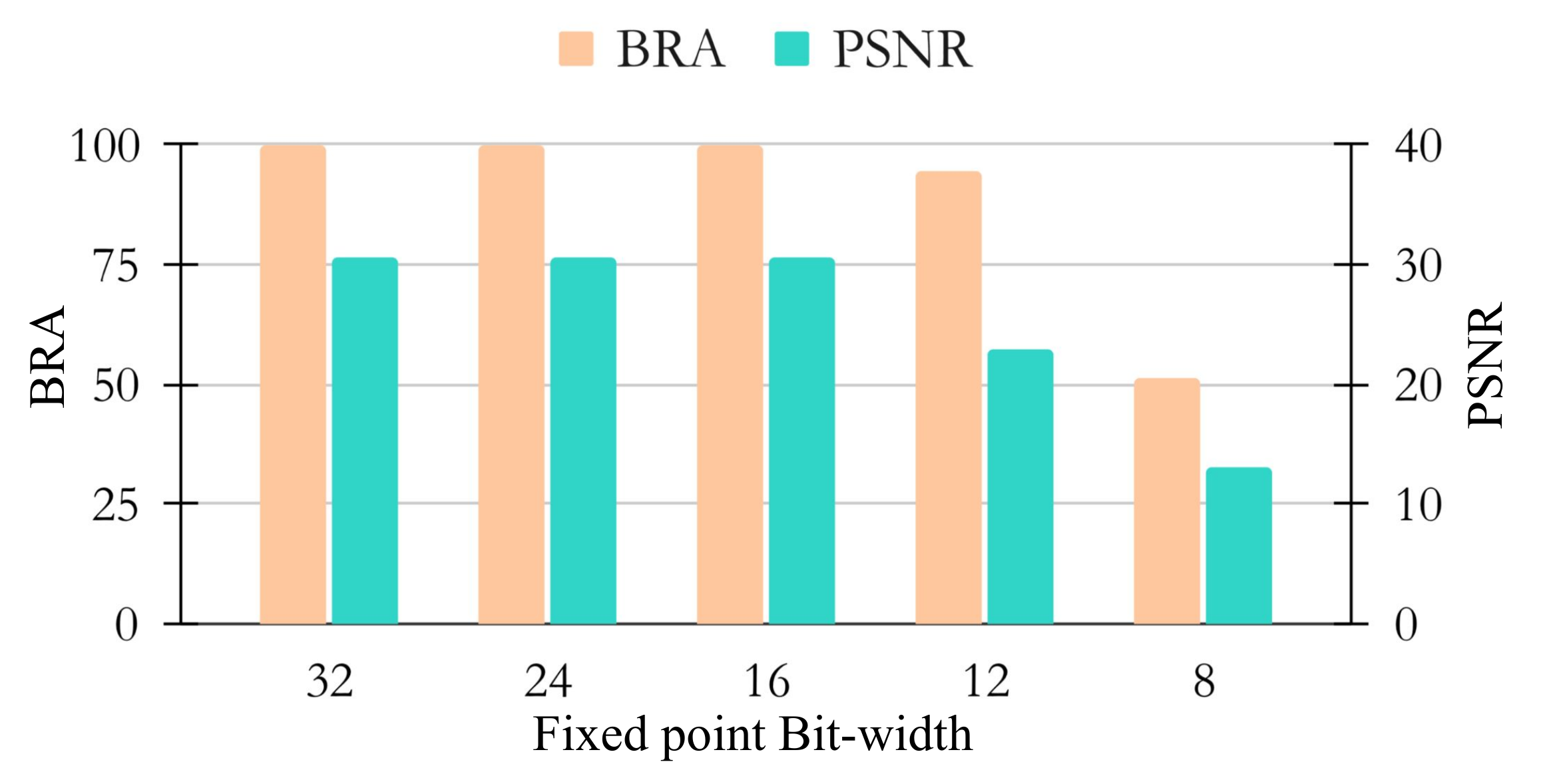}
\vspace{-6mm}
\caption{Watermarking success metrics for different fixed-point representations. 
A high value for both BRA and PSNR is desirable for accurate message recovery and imperceptibility. }
\label{fig:bitwidth}
\end{figure}

We implement the individual submodules described in Section~\ref{sec:implementation} using Vivado HLS for the Xilinx XCVU13P FPGA board. 
First, we perform a search over the bit-width of the fixed point representation of the network weights and intermediate outputs. 
Our goal is to find the lowest bit-width that does not compromise on message recovery and imperceptibility. 
Figure~\ref{fig:bitwidth} indicates the BRA and PSNR of different bit-width implementations of FastStamp. 
Based on this analysis, we use a 16-bit representation with 6 bits for the integer and 10 bits for the decimal representation. 
Next, we explore pipelining and loop unrolling options in our Vivado HLS implementation of various submodules.
Complete loop unrolling in submodule implementations was an infeasible design choice for FastStamp since it exceeded the available resources on the device.
We found that partial loop unrolling with factors of 8 and pipelining loops that were completely unrolled 
were the most effective optimizations for FastStamp.

\begin{figure}[htp]
\centering
\includegraphics[width=\linewidth]{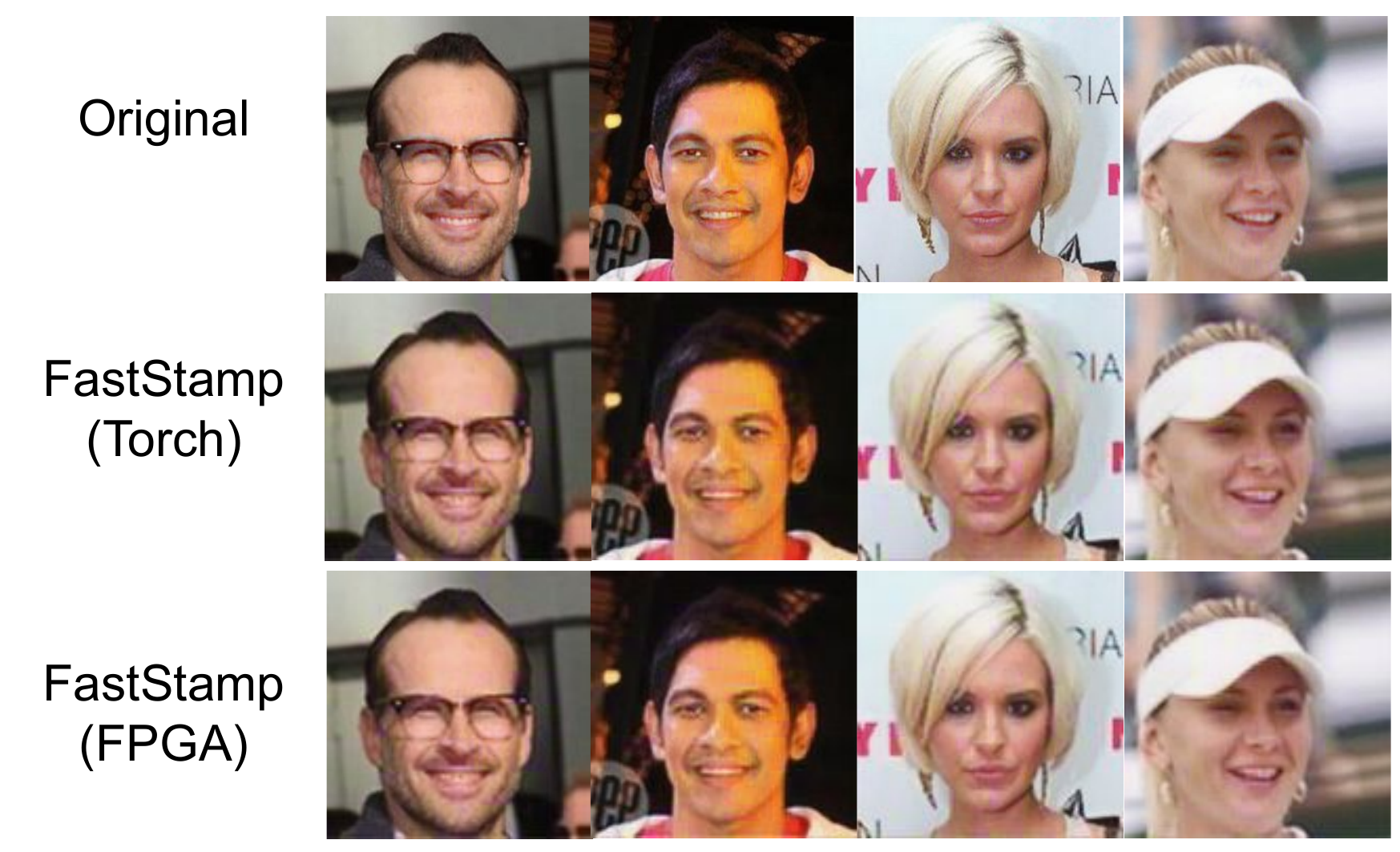}
\caption{Sample image outputs of FastStamp optimized design and PyTorch implementation with the original image}
\label{fig:sampleimage}
\end{figure}

While effective pipelining and loop unrolling resulted in a significant reduction in resource utilization, the LUT requirement of our design still exceeded the available resources on the device. 
Reducing the loop unrolling factor was an effective strategy to reduce LUT utilization but resulted in significant increases in encoding latency. 
To avoid latency increase, the next strategy we applied was distributing variables, such as model weights, that were initially all implemented as LUTs, to the BRAM on our board. 
We also force operations, such as multiplication in the separable convolution into DSP blocks, which results in lower LUT utilization.
We utilize per layer reuse factor to tune the inference latency versus utilization of FPGA resources and enable parallelization. 
This allows us to process multiple MAC operations at every unit of time. 
Using all of these optimizations, we are able to store our model entirely in the on-chip memory of the FPGA and perform low latency encoding while avoiding communication with off-chip memory. 
Table~\ref{tab:fpga} lists the resource utilization, performance, and correctness of some of the design choices that led to our most optimized design, \textit{FixedPoint-16-Optimized}. 
Figure~\ref{fig:sampleimage} shows sample outputs of this optimized FPGA implementation and compares them with the GPU implementation in PyTorch. 

\vspace{-3mm}
\subsection{\textbf{Performance and Power Analysis}}\label{sec:power}
Table~\ref{tab:cpugpu} compares the inference time and power requirement of our optimized FPGA implementation with the highly optimized CPU and GPU implementation of FastStamp and the open source implementations of prior neural watermarking systems. 
We benchmark the optimized PyTorch implementation of FastStamp on the Nvidia Tesla V100 GPU. 
The CPU implementation is a NumPy inference program written by us and optimized fully. 
We measure the power consumption for the GPU benchmarks using the Nvidia power measurement tool (\textit{Nvidia-smi}) running on \textit{Linux} operating system, which is invoked during program execution. 
For our FPGA implementations, we synthesize our designs using Xilinx Vivado 2020.1. 
We then integrate the synthesized modules accompanied by the corresponding peripherals into a system-level schematic using the Vivado IP Integrator. 
The frequency is set to 200\,MHz, and power consumption is estimated using the synthesis tool. 
Our FPGA implementation achieves ${\sim}68{\times}$ faster speed against prior work's GPU implementation and ${\sim}10{\times}$ faster speed against FastStamp's GPU implementation at a $3{\times}$ lower power requirement. 

\setlength\tabcolsep{7pt}
\begin{table}[htp]
\centering
\begin{tabular}{l|rr}
\multicolumn{1}{c|}{Implementation} &
\multicolumn{1}{|r}{Time (ms)} & 
\multicolumn{1}{r}{Power (W)} \\
\midrule
StegaStamp GPU~\cite{tancik2020stegastamp} & 205 & 76 \\
HiDDeN GPU~\cite{zhu2018hidden} & 234 & 65 \\
\midrule
FastStamp CPU  & 326 & --- \\
FastStamp GPU  & 30 & 59 \\
FastStamp FPGA & 3  &  19\\
\bottomrule
\multicolumn{1}{c}{}
\end{tabular}
\caption{Power consumption and wall-clock time (in milliseconds) required to generate a single watermarked image per implementation. 
}
\label{tab:cpugpu}
\end{table}

\vspace{-8mm}
\section{Conclusion}
We propose an efficient image watermarking model that matches or even outperforms prior neural image watermarking and steganography models while utilizing significantly fewer parameters. 
By leveraging an efficient secret message upsampling module, depthwise separable convolutions, and 2-D upsampling, we were able to train a smaller model while preserving success metrics for watermarking tasks. 
Finally, we implement our encoder model on an FPGA to achieve $68{\times}$ higher throughput as compared to prior GPU implementations. 
Our implementation allows watermark embedding directly at the hardware source, which not only secures the image capture and transmission pipeline but also reduces latency in embedding the watermark. 
In the process of this implementation, we develop reconfigurable sub-modules which can accelerate convolutional downsampling and upsampling networks on hardware.

\section{Acknowledgments}
This work was supported in part by National Science Foundation (NSF) Trust-Hub under award number CNS2016737, NSF TILOS under award number CCF-2112665, NSF Cooperative Agreement OAC-2117997 (\href{https://a3d3.ai}{a3d3.ai}), DoD UCR W911NF2020267 (MCA S-001364), and ARO MURI under award number W911NF-20-S-0009. 
Thanks to the Fast ML collective (\href{https://fastmachinelearning.org}{fastmachinelearning.org}) for support.

\bibliographystyle{ACM-Reference-Format}
\bibliography{main}

\end{document}